\documentclass{article}

\usepackage{neurips_2024}

\usepackage{wrapfig}

\usepackage{hyperref}

\usepackage{multirow}
\usepackage{pifont}
\usepackage{graphicx}
\usepackage[utf8]{inputenc} 
\usepackage[T1]{fontenc}    
\usepackage{hyperref}       
\usepackage{url}            
\usepackage{booktabs}       
\usepackage{amsfonts}       
\usepackage{nicefrac}       
\usepackage{microtype}      
\usepackage{xcolor}         

\title{\textit{RoboUniView}: Visual-Language Model with Unified View Representation for Robotic Manipulation}

\author{
    Fanfan Liu \\
    Meituan \\
    Beijing, China 100012 \\
    \texttt{liufanfan03@meituan.com}
    \And
    Feng Yan \\
    Meituan \\
    Beijing, China 100012 \\
    \texttt{yanfeng05@meituan.com} \\
    \And
    Liming Zheng \\
    Meituan \\
    Beijing, China 100012 \\
    \texttt{zhengliming04@meituan.com} \\
    \And
    Chengjian Feng \\
    Meituan \\
    Shenzhen, China 518110 \\
    \texttt{fengchengjian@meituan.com} \\
    \And
    Yiyang Huang \\
    Meituan \\
    Beijing, China 100012 \\
    \texttt{huangyiyang02@meituan.com} \\
    \And
    Lin Ma\thanks{Lin Ma is the corresponding author.} \\
    Meituan \\
    Beijing, China 100012 \\
    \texttt{malin11@meituan.com}
}

\begin{document}
\maketitle{}
\begin{abstract}

Utilizing Vision-Language Models (VLMs) for robotic manipulation represents a novel paradigm, aiming to enhance the model's ability to generalize to new objects and instructions. However, due to variations in camera specifications and mounting positions, existing methods exhibit significant performance disparities across different robotic platforms. To address this challenge, we propose \textit{RoboUniView} in this paper, an innovative approach that decouples visual feature extraction from action learning. We first learn a unified view representation from multi-perspective views by pre-training on readily accessible data, and then derive actions from this unified view representation to control robotic manipulation. This unified view representation more accurately mirrors the physical world and is not constrained by the robotic platform's camera parameters. Thanks to this methodology, we achieve state-of-the-art performance on the demanding CALVIN benchmark, enhancing the success rate in the $D \to D$ setting from 93.0\% to 96.2\%, and in the $ABC \to D$ setting from 92.2\% to 94.2\%. Moreover, our model exhibits outstanding adaptability and flexibility: it maintains high performance under unseen camera parameters, can utilize multiple datasets with varying camera parameters, and is capable of joint cross-task learning across datasets. Code is provided for re-implementation. \href{https://github.com/liufanfanlff/RoboUniview}{https://github.com/liufanfanlff/RoboUniview}

\end{abstract}

\section{Introduction}

Recent developments in foundation models \cite{radford2021learning, dosovitskiy2020image, vaswani2017attention} show significant advancements, demonstrating robust capabilities across a diverse array of tasks such as visual question answering (VQA) \cite{ zhou2022vlue}, open-vocabulary object detection and segmentation \cite{lin2022learning, gu2021open, ren2024grounded}, and comprehensive text-image understanding \cite{wei2023vary}. These achievements unequivocally motivate continued research into the effective integration of these models' capabilities into robotic control systems. 

Building on this momentum, the academic community diverge into two prominent methodologies: The first approach utilizes prompt tuning of out-of-the-box Large Language Models (LLMs) and Vision Language Models (VLMs) for zero-shot planning and task decomposition, which is subsequently complemented by the activation of a low-level skill library \cite{ahn2022can, driess2023palm,huang2022language,mu2024embodiedgpt,singh2023progprompt,wu2023tidybot}. These methods require intricate prompt tuning and strategic logic. The second approach focuses on imitation learning or reinforcement learning, using extensive robotic datasets \cite{brohan2022rt,brohan2023rt,li2023vision,ke20243d,black2023zero}. Drawing on the development trajectory of autonomous driving technology \cite{hu2023planning,xu2023drivegpt4,mao2023gpt,jiang2023vad}, there methods hold greater potential for achieving general intelligence in the field of robotic manipulation. However, existing methods exhibit significant performance disparities across different robotic platforms about the second approach. One of the main reasons is the differences in camera specifications and installation positions, which make it difficult for the models to accurately understand the real physical space from varied images, thus affecting the accuracy of their action predictions. We validate this with the state-of-the-art method, RoboFlamingo \cite{li2023vision}, and find that merely changing the camera parameters during inference cause a significant drop in success rate, from 86.3\% to 80.8\%. Although some methods are attempting to address this issue, such as RT-X \cite{linhartova2010rtx}, which trains models with more data collection, and 3D Diffusion Actor \cite{ke20243d}, which adds depth or point clouds to the input, but these approaches undoubtedly increase the workload and hardware costs.

\begin{figure}[t]
\centering
\includegraphics[width=1\textwidth]{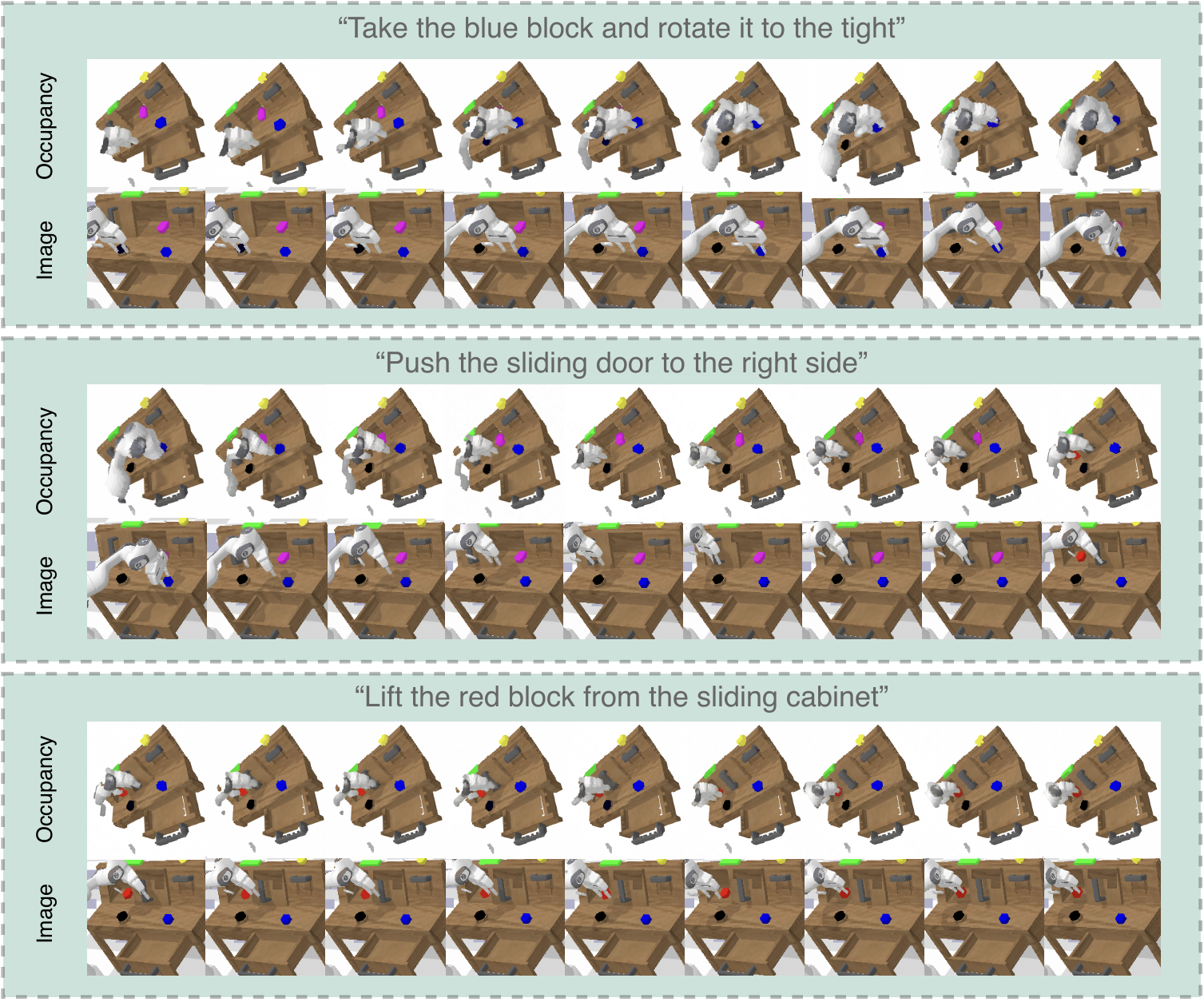}
\caption{The visualization of \textit{RoboUniView} on $D \to D$ split. The first row shows the predicted occupancy, and the second row shows the predicted rollouts.}
\label{fig:viz} 
\end{figure}

To address such limitations, this paper introduces \textit{RoboUniView}, a novel Visual-Language model with a unified view representation for robotic manipulation. Specifically, it decouples visual feature extraction from action learning. For visual feature extraction, it unifies the multi-perspective views into a unified view representation. To achieve this, we are inspired by BEVFormer \cite{li2022bevformer}  and propose a plug-and-play plugin called UVFormer, which can be integrated into any multi-modal model. This plugin is pre-trained on 3D occupancy task, taking multi-perspective views and corresponding camera parameters as inputs, and outputs the occupancy status and RGB values of each cell in the 3D grid, and gains a deeper understanding of the real physical world. It is noteworthy that our pre-training task only requires simple RGB-D images and does not need expensive manually annotated labels (such as semantic segmentation, objects, actions, etc.). For action learning, it directly outputs robotic actions from the unified view representation, following the design of OpenFlamingo \cite{awadalla2023openflamingo} and RoboFlamingo \cite{li2023vision}, which leverage publicly available pre-trained VLMs to integrate visual and language information.

A substantial amount of experimental evidence demonstrates that our model better comprehends the real physical world and significantly outperforms all existing methods in terms of performance. It also exhibits strong generalizability, maintaining high performance even in robots with unseen camera parameters. On the CALVIN \cite{mees2022calvin} dataset, a widely recognized simulation benchmark for long-horizon language-conditioned tasks, \textit{RoboUniView} establishes a new state-of-the-art by increasing the success rate from 88.7\% to 96.2\% in the $D \to D$ setting, from 82.4\% to 94.2\% in the $ABC \to D$ setting. Moreover, our model exhibits outstanding adaptability and flexibility: it maintains high performance under unseen camera parameters, can utilize multiple datasets with varying camera parameters, and is capable of joint cross-task learning across datasets. The visualizations are shown in the Figure \ref{fig:viz}, where \textit{RoboUniView} can capture the real physical environment and output effective actions.

To our knowledge, \textit{RoboUniView} represents the pioneering effort to demonstrate that a unified view representation, when coupled with pre-training on 3D occupancy task, significantly enhances both the performance and generalization capabilities of robotic manipulation across various camera parameters. The main contributions of this paper include: (1) Proposing a Visual-Language model with unified view representation for robotic manipulation, enhancing performance and generalization to robotic camera parameters. (2) proposing an effective pre-training method for obtaining a unified view representation to better comprehend the real physical world; (3) conducting extensive experiments to evaluate the performance of \textit{RoboUniView} in various settings, achieving state-of-the-art performance with a significant advantage on the CALVIN \cite{mees2022calvin} benchmark.

\section{Related Work}

\textbf{Language-conditioned Visuomotor Policies}. language-conditioned visuomotor robotic are the closest to human-like robotic operation, requiring robots to understand human commands and complete a variety of tasks based on visual feedback. This approach has a wide range of application prospects and challenges; therefore, in recent years, language-conditioned visuomotor policies have received considerable attention. Numerous works involve learning strategies and predictive models for robotic grasping \cite{gupta2018robot,zhou2023spil,zhang2022language,ding2019GCBC,mees2022HULC,lynch2020MCIL}. Robotic learning has seen rapid development in the areas of multi-tasking and language-conditioned learning, and \textit{RoboUniView} is built upon the foundation of these works.

\textbf{View Transform}.
View transformation has a wide range of applications in embodied AI, with the goal of converting perspective view features into a unified view aligned with the target task. In autonomous driving tasks, perception and planning need to be conducted from a Bird's Eye View (BEV) perspective. A direct method is to transform the perspective view into BEV using Inverse Perspective Mapping (IPM) \cite{reiher2020sim2real,can2021structured}. Additionally, Lift-Splat \cite{philion2020lift} generates BEV features based on depth distributions, while BEVFormer \cite{li2022bevformer} utilizes a Transformer to learn the conversion from perspective view features to BEV features. PETR \cite{liu2022petr} utilizes 3D positional encoding to implicitly complete view transformation. Subsequent work like UNIAD \cite{hu2023planning} completes end-to-end autonomous driving control based on View Transformation. In the field of robotic control, RVT \cite{goyal2023rvt} converts multiple perspective views into orthogonal perspective views, and VIHE \cite{wang2024vihe} transforms multiple perspective views into a hand-centric view. However, these works' viewpoint transformations rely on real depth and are still within a perspective view, leading to a mismatch between visual features and action space. In contrast, our work does not rely on depth maps and converts perspective view features into a unified 3D space view, aligning visual features with the action space.

\textbf{Generalization in robot learning}.
In the field of robotic learning, developing robotic controllers that can generalize across a variety of scenarios has been a long-standing research objective\cite{kaelbling2020foundation}
. Previously, some researchers have improved the generalization capabilities of robotic controllers by training models on a large and diverse datasets. For instance, tasks involving unseen objects \cite{young2021visual,levine2018learning,finn2017deep}, new combinations of objects and skills \cite{dasari2021transformers,jang2022bc}, new language instructions \cite{jiang2022vima,nair2022learning}, and new semantic object categories \cite{stone2023open}. Unlike these earlier efforts, recent works like RT2 \cite{brohan2023rt} and RoboFlamingo \cite{li2023vision} have utilized pre-trained models to enhance generalization capabilities, with these pre-trained models being trained on datasets much larger than those typically used for robots. Despite these advancements, current methods still focus on the capabilities for a single robotic embodiment. When the camera parameters of the robotic embodiment changes, the model non-functional. The goal of our method is to leverage the capabilities of pre-trained models while also being able to learn skills on  data from robots with different camera parameters, ultimately achieving generalization of a single model to different camera-configured robotic embodiment.

\textbf{Pre-training for robotic manipulation}.
Pre-training technology has driven the advancement of the entire artificial intelligence industry, ranging from early applications of various backbones pre-trained with supervised learning on ImageNet \cite{deng2009imagenet} for basic visual tasks, to large language models using self-supervised pre-training \cite{devlin2018bert}, and multi-modal models pre-trained with contrastive learning \cite{radford2021learning}.
In the field of robotic learning, previous researchers have also explored the use of pre-training techniques to enhance visual encoders \cite{karamcheti2023language}. Additionally, some work has employed pre-trained language models as instruction encoders or high-level planners \cite{ahn2022can, driess2023palm,huang2022language,mu2024embodiedgpt,singh2023progprompt,wu2023tidybot}. Recently, some researchers have used pre-training tasks related to robotic control to enhance the end-to-end capabilities of models, with RT2 \cite{brohan2023rt} and RoboFlamingo \cite{li2023vision} utilizing pre-trained VLMs. Unlike previous work, we not only utilize pre-trained VLM models, but we have also designed a novel pre-training approach to enhance the model's understanding of the real physical world.

\begin{figure}[t]
\centering
\includegraphics[width=1\textwidth]{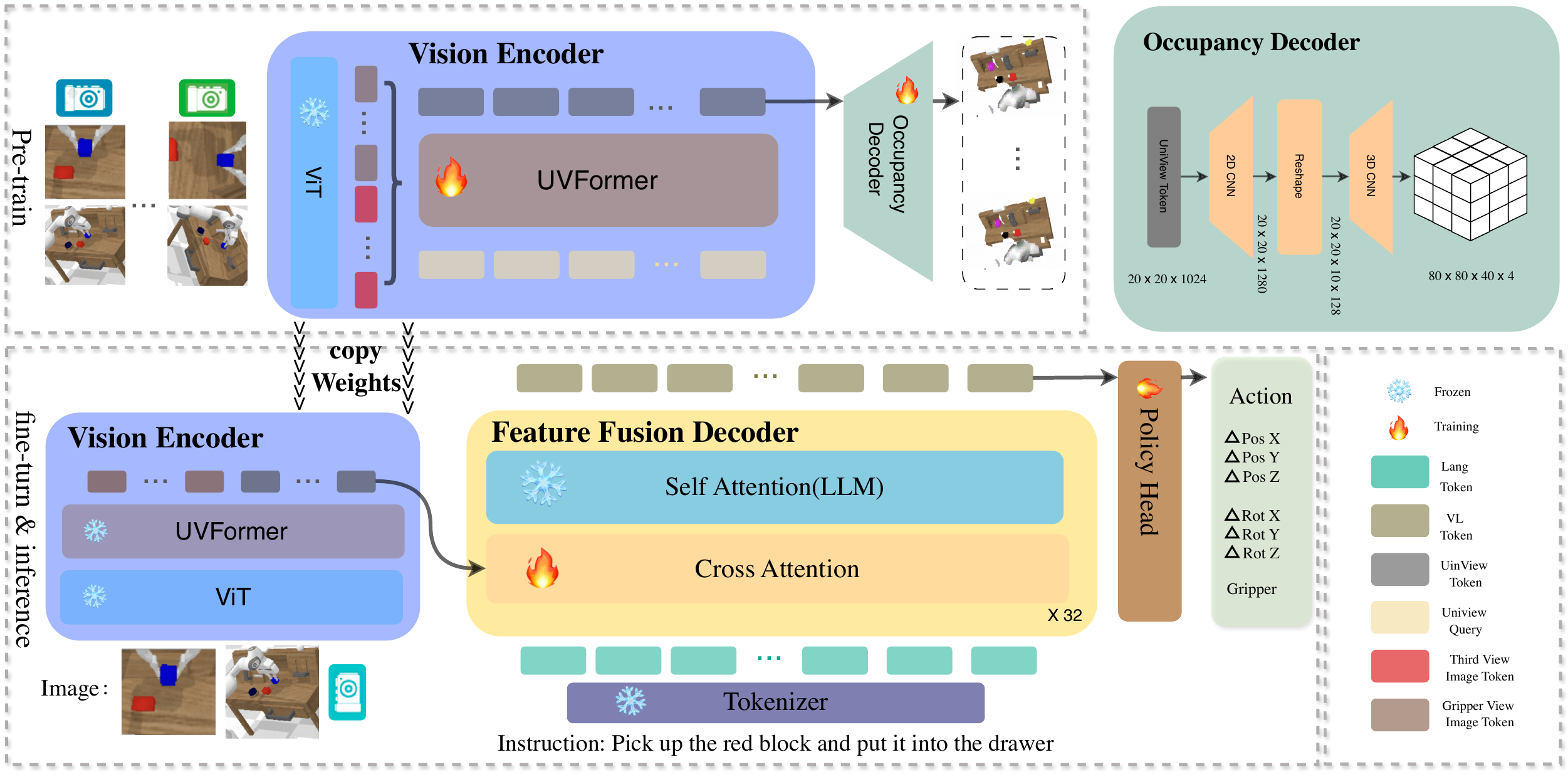}
\caption{Overview of the \textit{RoboUniView}. \textit{RoboUniView} is first pre-trained on the 3D occupancy task, and then fine-tuned on robot data to learn multi-task visual robot manipulation. }
\label{fig:robouniview} 
\end{figure}

\section{Method} 

The entire \textit{RoboUniView} framework is illustrated in Figure \ref{fig:robouniview}. During the forward process, multi-perspective images pass through Vision Encoder to extract wrist image features and the unified view representation. These are then combined with language tokens in the Feature Fusion Decoder to extract integrated vision-language features. Finally, these features pass through the policy head to execute robotic manipulation. The training process consists of two phases: during the pre-training phase, Vision Encoder undergoes training on a large dataset of easily accessible RGB-D images to learn robust unified view representation; during the fine-tuning phase, the model learns to predict robotic actions from the unified view representation, using paired images and action data.

\subsection{Architecture}

\subsubsection{Vision Encoder}

Our Vision Encoder comprises two parts: the Vision Transformer (ViT) and UVFormer. UVFormer's primary function is to transform multi-camera perspective view features, as outputted by the Vision Transformer (ViT), into unified view representation features.

\textbf{Vision Transformer}. 
We utilize pre-trained Vision Transformer (ViT) from CLIP \cite{radford2021learning} as the image backbone to extract features from images:
\begin{equation}
 X_{t}=\mathrm{ViT}\left(I_{t}\right),
\label{eq:vit}
\end{equation}
where $I_{t} \in \mathbb{R}^{N \times H \times W \times 3} $ and $X_{t} \in \mathbb{R}^{N \times H' \times W' \times C}$ represent the $N$ multi-perspective images and their corresponding extracted features at each time step $t$, respectively. The dimensions $ H \times W $ denote the size of the images, and $H' \times W'$ denote the dimensions of the output feature maps.

\begin{figure}[t]
\centering
\includegraphics[width=1\textwidth]{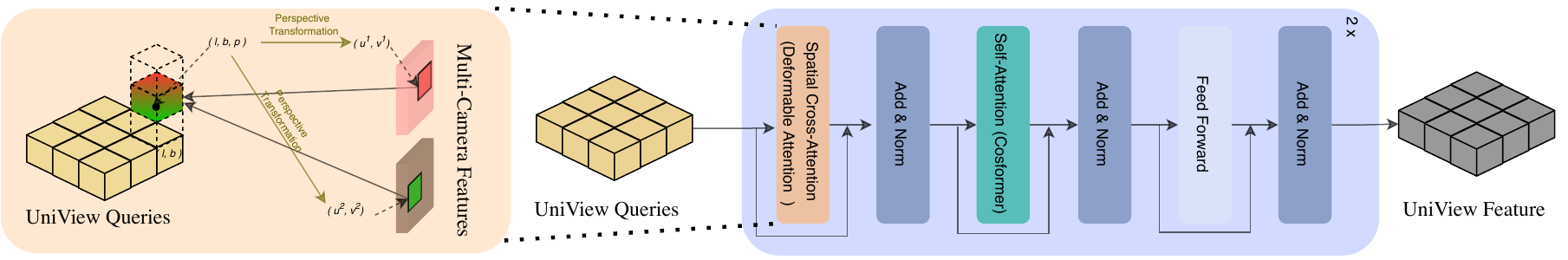}
\caption{UVFormer which contain grid-shaped UniView queries, Spatial Cross-Attention, and Self-Attention. Within the Spatial Cross-Attention, each UniView query interacts only with the image features of the pixel coordinates projected by its corresponding $P$ 3D points.}
\label{fig:UVfurmer}
\end{figure}

\textbf{UVFormer}. 
Inspired by BEVFormer\cite{li2022bevformer}, we design a simplified structure called UVFormer, which takes image features $X_{t}$, camera parameters $Cam$, and learnable UniView Queries $Q$ as inputs, and outputs a unified view representation $UF_{t}$. The process is defined by the equation:
\begin{equation}
UF_{t} = \mathrm{UVFormer}(Q, X_{t}, Cam).
\label{eq:2}
\end{equation}
$Q = \{Pos, Emb\}$, where $Pos \in \mathbb{R}^{L \times B \times 3P}$ and $Emb \in \mathbb{R}^{L \times B \times C}$ respectively represent the positions and learnable features of the queries, and $L, B, P$ define the spatial shape of the 3D grid located in the robot's operable space. Here, we set both L and B to be 20. Specifically, $Emb^{l,b} \in \mathbb{R}^{C}$ is responsible for the corresponding pillar cell region in the unified view space. Each pillar cell corresponds to size of $0.05^2$ meters in the real world, with $P$ 3D points uniformly distributed within 0.5-meter range along its vertical direction. $Cam$ represents the camera parameters of $N$ perspectives. $UF_{t} \in \mathbb{R}^{L \times B \times C}$ is the unified view representation, which encapsulates all relevant information across the $L \times B \times P$ 3D grid.

As shown in Figure \ref{fig:UVfurmer}, UVFormer includes two standard encoders. Each encoder layer consists of a Spatial Cross-Attention, a Self-Attention, and two Feed-Forward Networks (FFN). In the Spatial Cross-Attention layer, we utilize Deformable Attention ($\mathrm{DeformAttn}$) \cite{zhu2021deformable} due to the high computational cost associated with the standard multi-head attention mechanism. This resource-efficient attention layer allows each query $ Q^{l,b} $ to interact only with the features at specific positions under different camera perspectives, which are determined by the pixel positions resulting from the Perspective Transformation ($\mathrm{Proj}$) of the corresponding $P$ 3D points. The Spatial Cross-Attention (SCA) process is mathematically formulated as follows:
\begin{equation}
\mathrm{SCA}(Q^{l,b}, X_{t}, Cam) = \frac{1}{N}\sum_{n=1}^{N}\mathrm{DeformAttn}(Emb^{l,b}, \mathrm{Proj}(Pos^{l,b,P}, Cam^n), X_t^n).
\label{eq:SCA}
\end{equation}
In the Self-Attention layer, we implement an attention mechanism that omits the softmax function, providing performance that is comparable to traditional attention methods but with reduced resource consumption.

\subsubsection{Feature Fusion Decoder}

The Feature Fusion Decoder follows the design of OpenFlamingo\cite{awadalla2023openflamingo} and RoboFlamingo\cite{li2023vision}, taking $UF_{t}$, $X_t^{wrist}$, from the Vision Encoder, and language tokens $L_{t}\in\mathbb{R}^{I \times C} $ as inputs to generate vision-language features $VL_{t}\in\mathbb{R}^{I \times C}$. The entire structure includes $L$ layers of the decoder, each comprising a Cross Attention and Self Attention layer. Where the Self Attention is directly copied from pre-trained language models such as LLaMA\cite{touvron2023llama}, GPT-Neox\cite{black2022gpt}, or MPT\cite{team2023introducing}. The Cross Attention layer uses $L_{t}$ as queries and the concatenated $UF_{t}$ and $X_t^{wrist}$ as keys and values, and is fine-tuned on action data through imitation learning. Notably, $UF_{t}$ and $X_t^{wrist}$ are reshaped into $LB \times C$ and $H'W'\times C$ respectively. When features from the wrist perspective are missing in the data, they can be obtained through reconstruction to generate virtual $X_t^{wrist}$ from other perspectives. 

\subsubsection{Policy Head}

The Policy Head further transforms the output $VL_{t}$ into the pose of a 7-degree-of-freedom end-effector and the state of the gripper, which includes a Max-Pooling operation to aggregate information along the $I$ dimension, an LSTM\cite{hochreiter1997long} to integrate historical vision-language features, and MLP: 
\begin{equation}
a_t^{pose},a_t^{gripper}=\mathrm{MLP}(\mathrm{LSTM(MaxPooling(X_{VL}^{t}) ,h_{t-1})})
\label{eq:PolicyHead}
\end{equation}
where $h_{t-1}$ represents the hidden state at time $t$, and $a_t^{pose}$ and $a_t^{gripper}$ are the predicted pose of the end-effector and the state of the gripper, respectively.

\subsection{Training}

\subsubsection{Pre-training}
In the pre-training phase, we can use any data that includes RGB-D and corresponding camera parameters, rather than relying on expensive and high-quality datasets like RT-X\cite{padalkar2023open}. In this study, we initially adjust the robot platform's camera parameters within the CALVIN \cite{mees2022calvin} simulation environment to collect diverse RGB-D images. Subsequently, these images, along with their corresponding camera parameters, are used to generate RGB-enriched point clouds, which are then voxelized. This dataset is designated as \(Calvin_{rgbd}\).

As shown in Figure \ref{fig:robouniview}, to ensure that the unified view representation extracted by the Vision Encoder contains accurate physical information, we design a very simple occupancy decoder structure using pure convolution. This pre-training process takes multi-perspective RGB images and corresponding camera parameters as inputs, and outputs the occupancy and the RGB values of each grid cell. Here, we employ the L1 loss to supervise the RGB output of each grid cell and the cross-entropy loss to monitor the occupancy of the grid:
\begin{equation}
 l_{pre-train}= \lambda_{rgb}l_1^{rgb} +  l_{ce}^{occ},
\label{eq:occ_loss}
\end{equation}
where $\lambda_{rgb}$ corresponds to the weight, balancing RGB loss $l_1^{rgb}$ and occupancy loss $l_{ce}^{occ}$.

\subsubsection{Fine-tuning}

After pre-training, we obtain a robust unified view representation containing physical space information. To effectively control the robot, we simply fine-tune the Feature Fusion Decoder and Policy Head modules on multi-task grasping data to enable them to output specific actions. These actions include the position ($a_t^{pose}$ = \{$\Delta{pos}_t^x, \Delta{pos}_t^y, \Delta{pos}_t^z, \Delta{rot}_t^x, \Delta{rot}_t^y, \Delta{rot}_t^z\}$) of the 6-degree-of-freedom end-effector, as well as the gripper status $a_t^{gripper}$.

During this process, we use maximum likelihood imitation learning. Specifically, we use Mean Squared Error (MSE) loss to optimize the relative position and Binary Cross-Entropy (BCE) loss to optimize the gripper status :
\begin{equation}
l_a=\sum_t(\mathrm{MSE}(a_t^{pose},\hat{a}_t^{pose})+\lambda_{gripper}\mathrm{BCE}(a_t^{gripper},\hat{a}_t^{gripper})),
\label{eq:1}
\end{equation}
where $\lambda_{gripper}$ corresponds to the weight of gripper status loss, $\hat{a}_t^{pose}, \hat{a}_t^{gripper}$ is the demonstration for end-effector pose and gripper status at timestep $t$.

\section{Experiments} 

\subsection{Dataset } 
All our experiments are carried out using the language-conditioned CALVIN dataset \cite{mees2022calvin}, which encompasses 34 tasks within four distinct environments (A, B, C, and D). This dataset is developed on the PyBullet simulator and features manipulation scenarios involving a Franka Panda robotic arm. The environments differ in terms of table textures and object placements. CALVIN \cite{mees2022calvin} includes 24 hours of unstructured play data, of which 1\% is accompanied by language descriptions. Furthermore, CALVIN \cite{mees2022calvin} assesses 1,000 instruction sequences for sequential tasks, where each task requires the robot to sequentially execute five language instructions. The approach for each subsequent task is guided by a specific target instruction, allowing the agent to advance to the next objective only upon successful completion of the preceding task. 

\subsection{Comparisons with Other State-of-the-Art Methods}

\begin{table}[t]
  \centering
  \caption{\textit{RoboUniView} Benchmark Results. We compare average success rates between our model and prior benchmarks on multitask long-horizon control for 34 disparate tasks. A total of 1000 episodes are evaluated, each containing five consecutive tasks.}
  \label{tab:sota}
  \resizebox{\textwidth}{!}{%
  \begin{tabular}{ @{}c  c c c c c c c@{}}
    \toprule

    \multirow{2}{*}{Method} & \multirow{2}{*}{Train $\to$ Test}  &  \multicolumn{5}{c}{Task Completed in a Sequence } &   \multirow{2}{*}{Avg Len} \\
    \cline{3-7}
     &  &  1  &   2 &   3 &   4 &   5 &  \\
    
    \hline
    MCIL                        & $D \to D$ & 0.764 & 0.488 & 0.301 & 0.181 & 0.093 & 1.820\\
    GCBC                        & $D \to D$ & 0.647 & 0.284 & 0.122 & 0.049 & 0.013 & 1.110\\
    LCD                         & $D \to D$ & 0.887 & 0.699 & 0.545 & 0.427 & 0.322 & 2.880\\
    SPIL                        & $D \to D$ & 0.846 & 0.651 & 0.508 & 0.380 & 0.286 & 2.640\\
    HULC                        & $D \to D$ & 0.827 & 0.649 & 0.504 & 0.385 & 0.283 & 2.640\\
    RoboFlamingo                & $D \to D$ & 0.860 & 0.714 & 0.585 & 0.460 & 0.349 & 2.968\\
    HULC++                      & $D \to D$ & 0.930 & 0.790 & 0.640 & 0.520 & 0.400 & 3.3\\
    \midrule
    \textit{RoboUniView}(Ours)  & $D \to D$ & \textbf{0.962} & \textbf{0.888} & \textbf{0.776} & \textbf{0.666} & \textbf{0.563} & \textbf{3.855}\\
    \bottomrule
    MCIL                        & $ABC \to D$ & 0.304 & 0.013 & 0.002 & 0.000 & 0.000 & 0.400\\
    SPIL                        & $ABC \to D$ & 0.742 & 0.463 & 0.276 & 0.147 & 0.080 & 1.710\\
    HULC                        & $ABC \to D$ & 0.481 & 0.165 & 0.057 & 0.019 & 0.011 & 0.670\\
    RT-1                        & $ABC \to D$ & 0.533 & 0.222 & 0.094 & 0.038 & 0.013 & 0.900\\
    RoboFlamingo                & $ABC \to D$ & 0.824 & 0.619 & 0.466 & 0.331 & 0.235 & 2.470\\
    GR-1                        & $ABC \to D$ & 0.854 & 0.712 & 0.596 & 0.497 & 0.401 & 3.060\\
    3D Diffuser Actor           & $ABC \to D$ & 0.922 & 0.787 & 0.639 & 0.512 & 0.412 & 3.270\\
    \midrule
    \textit{RoboUniView}(Ours)  & $ABC \to D$ & \textbf{0.942} & \textbf{0.842} & \textbf{0.734} & \textbf{0.622} & \textbf{0.507} & \textbf{3.647}\\
    \bottomrule
  \end{tabular}}
\end{table}

\textbf{Imitation Performance}.
We fine-tune \textit{RoboUniView} using the demonstrations from the Split D training set, and evaluate its imitation performance on episodes sampled from Split D ($D \to D$). It takes about 2 days for training on 8 NVIDIA 80G A100 GPUs. As shown in Table \ref{tab:sota}, \textit{RoboUniView} significantly outperforms all methods across all metrics. The success rate of task1 is improved from 0.930 to 0.962. Even more impressive, in sequence of consecutive tasks, \textit{RoboUniView} increase the success rate of task5 from 0.400 to 0.563 and raise the average successful sequence length from 3.300 to 3.855. This result is particularly commendable as the complexity and challenge of subsequent tasks significantly increase with the progression of the tasks. This primarily stems from the fact that the initial state of each subsequent task is heavily dependent on the completion state of the previous task, leading to increasingly diverse starting conditions.

\textbf{Zero-Shot Generalization}. 
We also fine-tune \textit{RoboUniView} on the ABC split and test on the D split ($ABC \to D$), where the D split presents a completely different visual environment from ABC. It takes about 5 days for training on 8 NVIDIA 80G A100 GPUs.  As shown in Table \ref{tab:sota}, \textit{RoboUniView} improves the success rate of task1 from 0.922 to 0.942, and the average successful sequence length from 3.270 to 3.647, compare to best method. It demonstrate \textit{RoboUniView}'s strong capability in zero-shot generalization.

\subsection{Advanced Experiments}
\begin{figure}[t]
    \centering
    \includegraphics[width=1\textwidth]{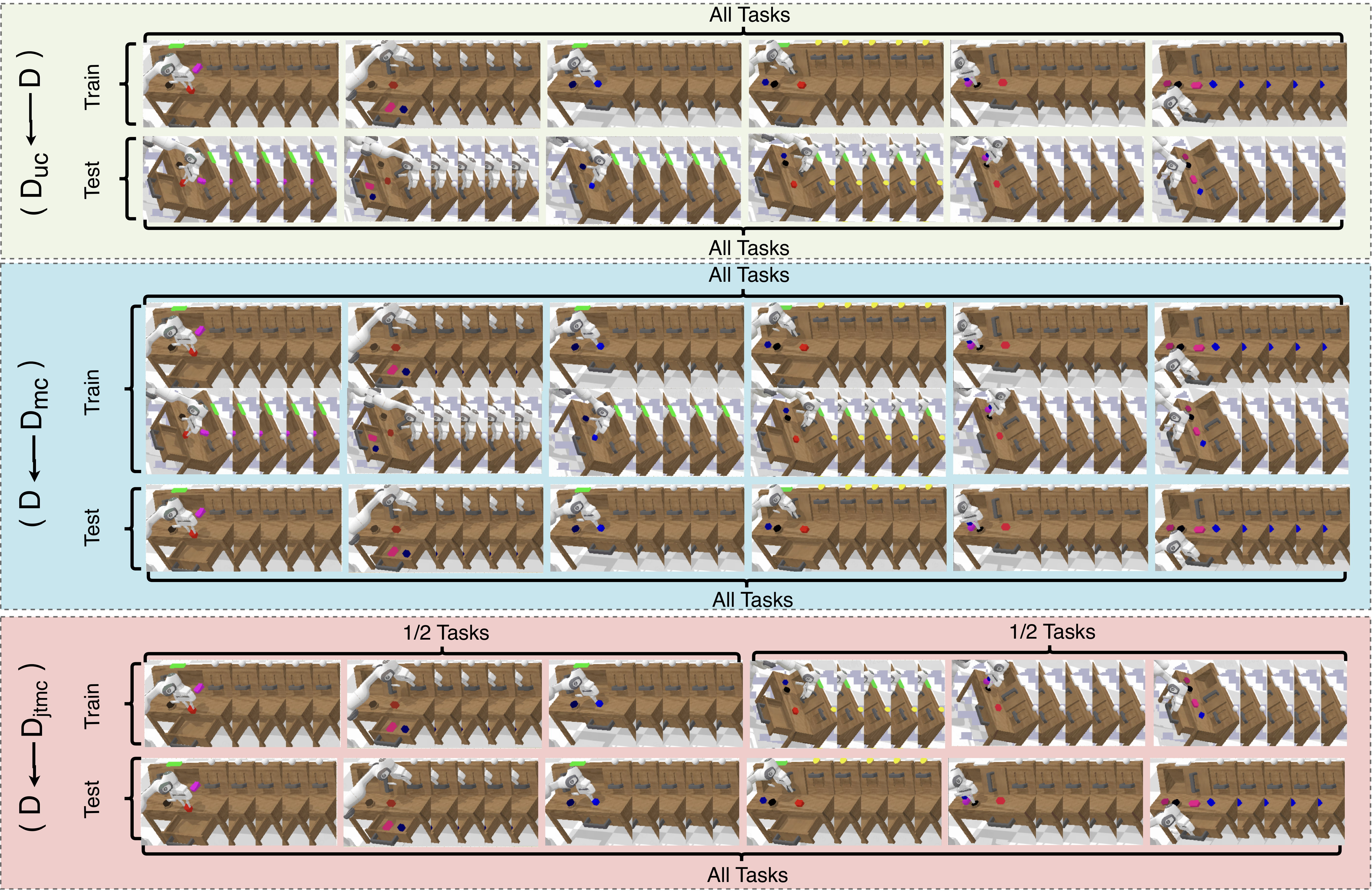}
    \caption{Visualization of environmental configurations in Advanced Experiments.}
    \label{fig:advanced} 
\end{figure}

To further validate the effectiveness of our method, we conduct three meaningful experiments using RoboFlamingo \cite{li2023vision} as the baseline.
(1) $D \to D_{uc}$: Training on the D split and testing on the D split with altered camera parameters.
(2) $D_{mc} \to D$: Training on the D split with two different sets of camera parameters, and testing on the D split.
(3) $D_{jtmc} \to D$: Training on the D split with two different sets of camera parameters, each set of which contain different tasks, and testing all tasks on the D split. It takes about 3 days for training on 8 NVIDIA 80G A100 GPUs
For more detailed information, please refer to Figure \ref{fig:advanced}.

\begin{table}[h]
  \centering
  \caption{Advanced experiments results.}
  \label{tab:AdvancedExperiments}
  \begin{tabular}{ @{}c c c@{\hspace{1pt}}c@{\hspace{1pt}}l @{\hspace{1pt}}c@{\hspace{8pt}} ccccc c@{}}
    \toprule
    & \multirow{2}{*}{Method}     & \multirow{2}{*}{Train} & \multirow{2}{*}{$\to$} & \multirow{2}{*}{Test}  && \multicolumn{5}{c}{Task Completed in a Sequence }  &   \multirow{2}{*}{Avg Len} \\
    \cline{7-11}
    &                             & &&                  && \multicolumn{1}{c}{1} & \multicolumn{1}{c}{2} & \multicolumn{1}{c}{3} & \multicolumn{1}{c}{4} & \multicolumn{1}{c}{5}  &\\
    \midrule
    A1 & Baseline                    & \multirow{2}{*}{$D$} & \multirow{2}{*}{$\to$} & \multirow{2}{*}{$D$}        && 0.860 & 0.714 & 0.585 & 0.460 & 0.349                 & 2.968 \\
    A2 & \textit{RoboUniView}  &                     &&            && 0.954 & 0.827 & 0.685 & 0.564 & 0.461                 & 3.491 \\
    \midrule
    B1 & Baseline                    & \multirow{2}{*}{$D$} & \multirow{2}{*}{$\to$} & \multirow{2}{*}{$D_{uc}$}   && 0.808 & 0.602 & 0.427 & 0.283 & 0.203                 & 2.323 \\
    B2 & \textit{RoboUniView}  &                     &&            && 0.956 & 0.825 & 0.681 & 0.561 & 0.460                 & 3.483 \\
    \midrule
    C1 & Baseline                    & \multirow{2}{*}{$D_{mc}$} & \multirow{2}{*}{$\to$} & \multirow{2}{*}{$D$}   && 0.821 & 0.706 & 0.568 & 0.441 & 0.325                 & 2.861 \\
    C2 & \textit{RoboUniView}  &                     &&            && 0.962 & 0.888 & 0.776 & 0.666 & 0.563                 & 3.855 \\
    \midrule
    D1 & Baseline                    & \multirow{2}{*}{$D_{jtmc}$} & \multirow{2}{*}{$\to$} & \multirow{2}{*}{$D$}  && 0.812 & 0.622 & 0.472 & 0.349 & 0.254                 & 2.509 \\
    D2 & \textit{RoboUniView}  &                     &&            && 0.966 & 0.859 & 0.703 & 0.590 & 0.473                 & 3.591 \\
    \bottomrule
  \end{tabular}
\end{table}

\textbf{Zero-Shot Unseen Camera Parameters Generalization}. 
To validate the generalization ability of \textit{RoboUniView} for zero-shot unseen camera parameters, we train both our model and the baseline on the D split, and subsequently test them on D split ($D \to D_{uc}$) with different camera parameters compare to training set. As illustrated in rows B1 and B2 of Table \ref{tab:AdvancedExperiments}, our method achieves high success rate of 0.956 in task1, significantly surpassing the baseline's 0.808. In comparison to rows A1 and A2, our method exhibits minimal fluctuation in success rate, less than 0.004 (0.681 vs 0.685) for all tasks, while the baseline method shows substantial variability, ranging from 0.052 (0.808 vs 0.860) to 0.177 (0.283 vs 0.460). This robust generalization capability is attributed to our unified view representation, which consistently maintains stable output. 

\textbf{Training on Multi Different Camera Parameters Datasets}. 
We re-render images based on D split training set with altered camera parameters, and merge them with the original D split training set to form the $D_{mc}$ training set. We train our model and the baseline model on the $D_{mc}$ set and test them on the D split set ($D_{mc} \to D$). As shown in rows C1 and C2 of Table \ref{tab:AdvancedExperiments}, compared to row A1, due to the new demonstration data being treated as interfering samples, the performance of the baseline method significantly declines (0.860 vs 0.821). In contrast, \textit{RoboUniView} achieve significant improvements (0.954 vs 0.962 for task1, 0.461 vs 0.563 for task5) compared to row A2. This demonstrates that training across multiple datasets to enhance performance becomes possibility.

\textbf{Joint Cross-task Learning for Cross-camera Parameters Generalization}.
The CALVIN \cite{mees2022calvin} dataset includes 34 tasks. We segment the D split training set into two subsets based on task type: $D_{train1}$ and $D_{train2}$, with each subset comprising 17 tasks. We change the camera parameters of $D_{train2}$, re-render the images based on demonstrations, and then merge them with $D_{train1}$ to form the $D_{jtmc}$ dataset. \textit{RoboUniView} and the baseline models are trained on $D_{jtmc}$ and test on all 34 tasks in the D split ($D_{jtmc} \to D$). As shown in rows D1 and D2 of Table \ref{tab:AdvancedExperiments}, thanks to our strategy of decoupling visual feature extraction from action learning, our model significantly outperforms the baseline model (+0.012 vs -0.48). This demonstrates that our model can engage in joint cross-task learning across datasets with different camera parameters, leveraging the unified view representation.

\subsection{Ablation Studies}

\begin{table}[h]
  \centering
  \caption{Ablation studies about UVFormer on the $D \to D$ dataset}
  \label{tab:uvformer}
 
  \begin{tabular}{@{}c l  c c c c c  c@{}}
    \toprule
        & \multirow{2}{*}{UVFormer} &  \multicolumn{5}{c}{Task Completed in a Sequence } &   \multirow{2}{*}{Avg Len} \\
    \cline{3-7}
        &                        &   \multicolumn{1}{c}{1} & \multicolumn{1}{c}{2} & \multicolumn{1}{c}{3} & \multicolumn{1}{c}{4} & \multicolumn{1}{c}{5}  & \\
    \midrule
    U1  & n/a (Baseline)                    & 0.860 & 0.714 & 0.585 & 0.460 & 0.349 & 2.968 \\
    U2  & Train (From Scratch)              & 0.893 & 0.763 & 0.613 & 0.491 & 0.408 & 3.168 \\
    U3  & +Pre-train+Fine-tune              & 0.912 & 0.771 & 0.622 & 0.503 & 0.395 & 3.203\\
    U4  & +Pre-train+Fine-tune (Frozen)     & 0.954 & 0.827 & 0.685 & 0.564 & 0.461 & 3.491 \\
    \bottomrule
  \end{tabular}
\end{table}
\textbf{UVFormer}.
UVFormer is the most critical component of our model, fundamentally determining the expression capability of the unified view representation. As shown in Table \ref{tab:uvformer}, we conduct thorough research: the first row U1 displays the baseline test results without UVFormer, the second row U2 shows the results after training our model directly on the $D \to D$ dataset; the third row U3 shows the results after pre-training UVFormer on the $Calvin_{rgbd}$ dataset followed by fine-tuning; the fourth row U4 shows the results when UVFormer is frozen during the fine-tuning process. Compared to the U1 baseline, the success rate of U2 increase from 0.860 to 0.893, demonstrating the importance of the UVFormer module and unified view representation. The success rates of U3 and U4 progressively increase to 0.954. These results indicate that the optimal performance of the model can be achieved by pre-training and freezing the UVFormer parameters during fine-tuning.

\begin{table}[h]
  \centering
  \caption{Ablation studies on different resolutions of UniView Queries on the $D \to D$ dataset}
  \label{tab:DifferentResolution}
  \begin{tabular}{ @{}c  c c c c c  c c@{}}
    \toprule
    \multirow{2}{*}{Resolution(l, b, p)}   &  \multicolumn{5}{c}{Task Completed in a Sequence } &   {\multirow{2}{*}{Avg Len}} & \multirow{2}{*}{Latency (ms)} \\  
    \cline{2-6}
     &  {1}  &    {2} &   {3} &   {4} &  {5}  &  & \\
    \midrule
    (0.200, 0.200, 0.250)   & 0.865 & 0.710 & 0.560 & 0.412 & 0.314 & 2.861 & 98\\
    (0.100, 0.100, 0.100)   & 0.902 & 0.761 & 0.614 & 0.481 & 0.381 & 3.139 & 101\\
    (0.050, 0.050, 0.100)   & 0.893 & 0.763 & 0.613 & 0.491 & 0.408 & 3.168 & 105\\
    (0.025, 0.025, 0.050)   & 0.906 & 0.760 & 0.615 & 0.491 & 0.412 & 3.184 & 205\\

    \bottomrule
  \end{tabular}
\end{table}
\textbf{Resolution of UniView Queries}. 
We adjust the resolution of UniView Queries to explore its impact on model performance and speed. As shown in Table \ref{tab:DifferentResolution}, with the resolution of UniView Queries increases, the performance of the model improves, but the latency of the model also increases. At resolutions below (0.050, 0.050, 0.100), the increase in resolution has minimal effect on latency. However, when the resolution increase to (0.025, 0.025, 0.050), the latency nearly doubles. To balance the performance and speed of the model, we ultimately chose a resolution of (0.050, 0.050, 0.100). Casually speaking, the baseline latency is 91 milliseconds (ms).
.

\section{Limitations and Future Work}
Despite the introduction of the unified view representation, which enhances the robustness to camera parameters, this paper still faces some limitations. 
Camera Calibration Dependency: Our approach relies on precise camera calibration. This dependency means that any slight misalignment or calibration error could significantly affect the accuracy and reliability of the model's output. Fortunately, camera calibration is a well-established and straightforward process, often involving the use of checkerboard or QR code patterns for calibration.
Simulation-Based Testing: Due to the lack of real-robot data, this paper is not able to deploy on real-world robots, which is also the direction of our future work. It is promising to note that, with the exponential growth of robotic data, we are optimistic that future research will showcase \textit{RoboUniView}'s effectiveness in real-world tasks.

\section{Conclusion}

This paper introduces \textit{RoboUniView}, a novel visual-language model with a unified view representation for robotic manipulation, which proposes the pre-training method for this unified view representation. \textit{RoboUniView} achieves state-of-the-art performance on benchmark datasets. Moreover, various experiments demonstrate our method's significant generalization advantages with data from different camera parameters. These strengths clearly pave the way for our next objective: training a comprehensive embodied intelligence model on diverse robotic datasets to master all skills and deploy on various platforms.





\end{document}